# CardioTabNet: A Novel Hybrid Transformer Model for Heart Disease Prediction using Tabular Medical Data

Md. Shaheenur Islam Sumon[1], Md. Sakib Bin Islam[1], Md. Sohanur Rahman[2], Md. Sakib Abrar Hossain[1], Amith Khandakar[2], Anwarul Hasan[3], M Murugappan[4,5*], Muhammad E. H. Chowdhury[2*]

[1]Department of Electrical Engineering, Qatar University, Doha 2713, Qatar.
[2]Department of Electrical and Electronic Engineering, University of Dhaka, Dhaka 1000, Bangladesh.
[3]Department of Industrial and Mechanical Engineering, Qatar University, Doha-2713, Qatar.
[4]Intelligent Signal Processing (ISP) Research Lab, Department of Electronics and Communication Engineering, Kuwait College of Science and Technology, Block 4, Doha, 13133, Kuwait.
[5]Department of Electronics and Communication Engineering, Vels Institute of Sciences, Technology, and Advanced Studies, Chennai, Tamilnadu, India.
*Correspondence: Muhammad E. H. Chowdhury (mchowdhury@qu.edu.qa), M Murugappan (m.murugappan@kcst.edu.kw)

**Abstract**
The early detection and prediction of cardiovascular diseases are crucial for reducing the severe morbidity and mortality associated with these conditions worldwide. A multi-headed self-attention mechanism, widely used in natural language processing (NLP), is operated by Transformers to understand feature interactions in feature spaces. However, the relationships between various features within biological systems remain ambiguous in these spaces, highlighting the necessity of early detection and prediction of cardiovascular diseases to reduce the severe morbidity and mortality with these conditions worldwide. We handle this issue with CardioTabNet, which exploits the strength of tab transformer to extract feature space which carries strong understanding of clinical cardiovascular data and its feature ranking. As a result, performance of downstream classical models significantly showed outstanding result. Our study utilizes the open-source dataset for heart disease prediction with 1190 instances and 11 features. In total, 11 features are divided into numerical (age, resting blood pressure, cholesterol, maximum heart rate, old peak, weight, and fasting blood sugar) and categorical (resting ECG, exercise angina, and ST slope). Tab transformer was used to extract important features and ranked them using random forest (RF) feature ranking algorithm. Ten machine-learning models were used to predict heart disease using selected features. After extracting high-quality features, the top downstream model (a hyper-tuned ExtraTree classifier) achieved an average accuracy rate of 94.1% and an average Area Under Curve (AUC) of 95.0%. Furthermore, a nomogram analysis was conducted to evaluate the model's effectiveness in cardiovascular risk assessment. A benchmarking study was conducted using state-of-the-art models to evaluate our transformer-driven framework.

## 1. Introduction

Heart diseases are among the leading causes of death and disability worldwide, affecting millions of people every year. According to the World Health Organization (WHO), 16% of all fatalities in 2019 [1] were attributed to cardiovascular diseases (CVD). Detecting and forecasting cardiac diseases at an early stage can help prevent or delay complications such as myocardial infarction and arrhythmia. However, the early diagnosis and prediction of heart diseases require numerous variables like age, gender, blood pressure, cholesterol levels, diabetes, smoking habits, family history, and lifestyle, is a challenging task [2-4]. Moreover, the presence of diverse risk factors and symptoms across various forms of heart disease complicates the process of determining the most suitable diagnostic test or treatment for a given patient. A comprehensive evaluation of a patient's cardiac status is crucial for the diagnosis and management of cardiovascular disorders. Through the utilization of computed tomography (CT), magnetic resonance imaging (MRI), electrocardiograms (ECGs), a solid assessment can be achieved [5]. However, due to resource limitations, testing with these diagnostic modalities may prove challenging to administer and sometimes less accurate Therefore, alternative and innovative methods for early prediction of cardiac diseases could save millions of lives, especially in developing nations [6, 7].

Detecting CVD early and intervening promptly is crucial for reducing premature death [8, 9]. Predictive models play a vital role in identifying high-risk patients, facilitating timely clinical interventions. Conditions such as coronary heart disease, heart failure, congenital heart disease, cyanotic heart disease, and cardiomyopathy are some of the many forms of CVD that affect health and well-being [10]. Coronary heart disease limits the supply of vital nutrients and oxygen to the heart by obstructing the coronary arteries. This often leads to potentially fatal outcomes. Heart failure, a condition in which the heart is unable to pump blood effectively, represents a more advanced stage of CVD, frequently triggered by factors such as coronary artery disease. Congenital heart disease is characterized by structural abnormalities in the heart, such as septal defects. It develops during fetal growth and are present from birth [11]. Cyanotic heart disease arises from restrictive defects that limit oxygen delivery or impede the blood flow [12]. Cardiomyopathy weakens the heart's ability to circulate blood, potentially leading to heart failure.

One potential approach to address this problem is the application of machine learning (ML) methodologies to analyze extensive and complex datasets related to cardiovascular diseases, to extract valuable insights and accurate predictions. ML, a subfield of artificial intelligence, enables computers to learn from data and perform tasks that is dependent on human intelligence [13, 14]. It has been extensively explored across various domains of biomedical research and healthcare, including genomics, proteomics, drug discovery, diagnosis, prognosis, and personalized medicine [8]. ML has shown significant promise, particularly in predicting CVD using diverse data sources, such as genomic information, clinical records, demographic data, lifestyle factors, biomarkers, ECG signals, and imaging data [15-17]. It can aid in identifying risk factors, patients' classification, estimation of disease development likelihood, and suggest appropriate actions or treatments. Despite all these advantages, ML faces several challenges and limitations in the context of CVD prediction. One significant challenge is handling the diversity and complexity of the data sources and formats. Datasets often vary in properties, sizes, distributions, quality, reliability coupled with containing errors, missing values, outliers, or noise. Additionally, some datasets may present multicollinearity or high dimensionality. These issues can affect the performance and generalizability of ML models [18]. Selecting the optimal algorithm for a given task depends on several factors, including the nature and size of the data, the computational cost and efficiency of the algorithm, and the complexity and interpretability of the model [19].

Compared to previous research that focused on heart disease datasets and achieved satisfactory results but did not explore robust datasets like ours or the use of embedding for categorical variables, we aim to evaluate the efficacy of the tab transformer architecture. Our hypothesis is that this architecture will lead to improved performance, as recent studies have suggested. A significant enhancement in our approach is the incorporation of tab transformer, a sophisticated deep-learning architecture designed for modeling tabular data. This innovative model transforms categorical feature embeddings into robust contextual embeddings, addressing a crucial aspect often overlooked in earlier methodologies. CardioTabNet strategically leverages this novel technique to significantly improve prediction accuracy, distinguishing it from previous models and providing a more comprehensive representation of data in the context of heart disease prediction.

Innovative predictive models that provide reliable, interpretable, and accurate cardiovascular risk assessments are urgently needed to address these challenges. The recent advances in ML, particularly in handling tabular medical data, offer a promising pathway to improve heart disease prediction. The objective of our study is to bridge this gap by proposing a novel model that utilizes state-of-the-art ML techniques to enhance predictive accuracy and provide insights into CVD risk factors.

A pioneering approach to predicting CVD based on the IEEE Data Port Heart Disease dataset [4, 10, 12, 22-24]. The motivation for this research is the growing global burden of cardiovascular diseases, which are the leading cause of death in the world. It is crucial to make accurate and timely predictions of CVD in order to reduce mortality rates and initiate early interventions. There is a clear need for more robust, interpretable, and efficient systems that can use modern ML techniques and large datasets to improve prediction accuracy and clinical applicability despite the availability of numerous predictive models. In summary, we have made the following contributions:

- In this study, we present CardioTabNet, a state-of-the-art model that was specifically designed to predict CVDs. A rigorous and comprehensive statistical analysis was conducted to provide a solid foundation for the findings of this study. This model incorporates the rich set of data available from IEEE Data Port to enhance the accuracy and reliability of cardiovascular risk assessment. Our proposed model, CardioTabNet, is validated and credible through this analysis.
- CardioTabNet utilizes tab transformer to extract essential features from a dataset. A feature ranking strategy was used to identify the top 10 features, which were then used to train 10 classical machine-learning models. In this way, a more focused and efficient predictive model can be developed.
- There has been a notable improvement in experimental results compared to previous studies in this area. CardioTabNet's effectiveness combined with the top 10 features has resulted in superior predictive performance for CVD.
- We incorporated logistic regression and nomogram analysis to further enhance binary classification precision. The sophisticated analysis enhances accuracy and introduces a scoring system that facilitates nuanced differentiation between positive and negative categories.

By following this structured format, our research establishes a robust foundation for advancing the field of prediction of CVD, displaying the efficacy of CardioTabNet and contributing valuable insights for future studies. The subsequent sections of this manuscript are structured as follows: A review of related works is presented in Section 2. The comprehensive methodology is provided in section 3, while the results and discussion are presented in section 4. The study concludes in Section 5 with a formal summary and future considerations.

## 2. Related Works

This section provides an overview of several recent studies that utilize ML techniques and the IEEE Data Port dataset to predict cardiac disease. Five distinct datasets from various sources are merged to form the IEEE Data Port dataset; these include the Hungarian, Cleveland, Long Beach, Virginia, Switzerland, and Statlog datasets. The dataset includes 1190 instances and 11 features, which include both numeric and nominal attributes. The target variable denotes the presence or absence of cardiac disease in patients as a binary class.

Various ML models have been applied to the prediction of heart disease in recent research, with notable success. In Tiwari et al.'s study [4], classifiers such as ExtraTrees, Random Forest, and XGBoost were applied to the IEEE Data Port and achieved high accuracy (92.34%) and F1-scores of 92.74%, showing robust recall (93.49%) and sensitivity (91.07%). Similar to that study, another study [5] evaluated multiple ML models using the IEEE Data Port, including Logistic Regression, KNN, Decision Tree, and Random Forest, resulting in an accuracy score of 91.60%. A recall of 94.30%, sensitivity of 88.39%, and an F1-score of 92.43% were reported in the study. The comprehensive investigation highlighted the effectiveness of ensemble methods as a means of predicting heart disease, reinforcing the notion that combining different models can lead to more accurate outcomes. Nagarajan et al. [6] contributed further to this field by employing Logistic Regression, KNN, and Random Forest models, which resulted in an accuracy of 90.67%, sensitivity of 92.68%, and specificity of 88.39%. As a result of their work, they were able to demonstrate that combinations of models can achieve balanced and comprehensive performance in classification tasks. A different approach [7] integrated SHAP (SHapley Additive Explanations) for feature importance analysis and XGBoost for classification. By utilizing the IEEE Data Port, this study sought to increase interpretability while achieving a respectable accuracy of 90.08%, with a sensitivity of 91.46% and specificity of 88.39%.

A number of datasets, including the IEEE Data Port and the Mendeley Data Center cardiovascular data, were used by Doppala et al. [8] to illustrate the versatility of ML. Across different datasets, they found that Naive Bayes, Random Forest, and XGBoost consistently performed well, achieving impressive accuracy levels of 96.75% for the Mendeley dataset and 93.39% for IEEE Data Port. There were also traditional models [9] that showed competitive performance. As an example, using Decision Trees, Random Forest, and SVM with IEEE Data Port dataset resulted in an accuracy of 85.12%. The results show that classical models can still provide reliable results even though newer techniques offer enhancements. According to Paul et al. [10], it is crucial to account for dataset variability. They used artificial neural networks with scaled conjugate gradient backpropagation to analyze the Cleveland Hungarian Statlog dataset and the Cleveland processed heart dataset. Consequently, the accuracy of the model varied significantly, ranging from 63.38% to 88.48%, emphasizing the importance of the choice of dataset. Advanced techniques were further explored by Baccouche et al. [11], who combined BiLSTM or BiGRU models with CNN to form an ensemble classifier. Based on their study of the MIT-BIH Arrhythmia Database, they achieved excellent accuracy and F1 scores ranging from 91% to 96%, demonstrating the potential of deep learning in medical applications. Similarly, Dubey et al. [12] used the IEEE CHD dataset, applying Naive Bayes, Bayes Nets, and Multilayer Perceptrons, and achieved 93.67% accuracy. In light of this, Bayesian methods are a reliable method for predicting heart disease provided they are applied to medical data. According to Mohan et al. [13] and Mahmud et al. [14], using the UCI Dataset, different models produced different results. Mahmud et al. reached 85.71% accuracy using Decision Trees and Neural Networks while Mohan et al. achieved 88.7% accuracy using Decision Trees and Neural Networks. As a result of these findings, models must be carefully selected in accordance with the specific dataset and research objectives.

Moreover, studies using Kaggle and Cleveland Clinic Foundation datasets indicated that KNN often outperformed other models. In contrast, weighted Naive Bayes and XGBoost with DBSCAN and SMOTE-ENN [15] excelled in precision, demonstrating the strengths of probabilistic methods in classification tasks. Using a MultiLayer Perceptron, an accuracy of 87.28% and a high AUC score of 0.95 were demonstrated, highlighting the benefits of neural networks in the application domain.

An overview of the prior research that utilized this dataset to forecast heart disease is presented in Table 1. According to Table 1, we developed our research questions after reviewing the most recent and prominent research in this field.

**Table 1** presents a compilation of previous research endeavors focused on the prediction of cardiac disease

| Author Name | Dataset | Methods | Results and Observation |
| --- | --- | --- | --- |
| **Tiwari et al. [20]** | IEEE Data Port | ExtraTrees Classifier, Random Forest, XGBoost, GBM and Logistic Regression | Accuracy: 92.34%, Recall: 93.49%, Sensitivity: 91.07%, F1-Score: 92.74% |
| **Rajdhan et al. [21]** | IEEE Data Port | Logistic Regression, KNN, Decision Tree, Random Forest, Support Vector Machine (SVM), Naive Bayes, XGBoost, LightGBM, CatBoost | Accuracy: 91.60%, Recall: 94.30%, Sensitivity: 88.39%, F1-Score: 92.43% |
| **Nagarajan et al. [22]** | IEEE Data Port | Logistic Regression, KNN, Decision Tree, Random Forest, SVM, Naive Bayes, XGBoost | Accuracy: 90.67%, Sensitivity: 92.68%, Specificity: 88.39%, F1-score: 91.18% |
| **Tjoa et al. [23]** | IEEE Data Port | SHAP (SHapley Additive exPlanations) framework for feature importance analysis and XGBoost for classification | Accuracy: 90.08%, Sensitivity: 91.46%, Specificity: 88.39%, F1-score: 90.43% |
| **Doppala et al. [24]** | IEEE Data Port | Naive Bayes, Random Forest, XGBoost, SVM | Mendeley Data Center cardiovascular disease dataset: 96.75% accuracy, IEEE |

| | | | Data Port: 93.39% accuracy, and 88.24% accuracy on the Cleveland dataset. |
|---|---|---|---|
| **Dinesh et al. [25]** | IEEE Data Port | Decision Tree (DT), Random Forest (RF), Support Vector Machine (SVM), K-Nearest Neighbors (KNN), and Logistic Regression. | Accuracy: 85.12% |
| **Paul et al. [26]** | Cleveland Hungarian Statlog heart dataset, Cleveland processed heart dataset | Scaled conjugate gradient backpropagation in artificial neural networks | Minimum accuracy is 63.3803% for the Cleveland processed heart dataset and 88.4754% for the Cleveland Hungarian Statlog heart dataset |
| **Baccouche et al. [27]** | MIT-BIH Arrhythmia Database | Ensemble classifier with BiLSTM or BiGRU model with CNN model | Accuracy and F1-score between 91% and 96% |
| **Dubey et al. [28]** | IEEE CHD | Naive Bayes, Bayes Net, and Multilayer Perceptron | Accuracy: 93.67% |
| **Mohan et al. [29]** | UCI Dataset | Decision Trees (DT), Neural Networks (NN), Support Vector Machines (SVM), and K-closest Neighbors (KNN). | Accuracy: 88.7% |
| **Mahmud et al. [33]** | UCI Dataset | Logistic Regression, Decision Tree, Support Vector Machine, etc. | Accuracy: 85.71% |
| **Sharma et al. [30]** | UCI Dataset | Random Forest, SVM, Naive Bayes, and Decision Tree ML techniques. | The Naïve Bayes with a 90% accuracy rate, and Random Forest with an accuracy of 87% |
| **Moreno-Sanchez et al. [31]** | Heart Failure Survival Dataset | SCI-XAI automated data processing pipeline | Balanced Accuracy of 0.74 (std 0.03). |
| **Sarra et al. [32]** | Cleveland and Statlog (heart) datasets. | The chi-squared-SVM technique | The suggested model improved accuracy from 85.29 to 89.7%. |
| **Reddy et al. [33]** | Cleveland, Switzerland, Hungarian, V.A. Medical, and Statlog project heart disease datasets | SVM-linear, Naive Bayes, and Neural Network. | Best Accuracy at 84.81%. |
| **An Dinh et al.[34]** | National Health and Nutrition Examination Survey (NHANES) Dataset | Information gain of tree-based models identifies patient data characteristics | WEM performs best with an AU-ROC score of 83.1% without laboratory data and 83.9% with lab data. |
| **Shah et al. [35]** | Cleveland database of UCI repository of heart disease patients | Naive Bayes, decision tree, K-nearest neighbor, and random forest algorithm | K-nearest neighbor scored best in accuracy. |
| **Nagavelli et al. [36]** | UCI repository and Kaggle datasets | Weighted Naive Bayes, XGBoost, duality optimization, and XGBoost with DBSCAN and SMOTE-ENN. | XGBoost with DBSCAN and SMOTE-ENN had the greatest precision, accuracy, f1-measure, and recall. |
| **Bhatt et al. [37]** | Cleveland Clinic Foundation | MultiLayer Perceptron (MLP) | MLP has cross-validation accuracy of 87.28%, recall, precision, F1 score, and |

|  |  |  |  | AUC scores of 84.85, 88.70, 86.71, and 0.95. |
|---|---|---|---|---|
|  | **Ours** | IEEE Data Port | CardioTabNet | Accuracy: 94.08%, Precision: 92.84%, Recall: 97.37%, F1 Score: 94.47%, Specificity: 91.92% |

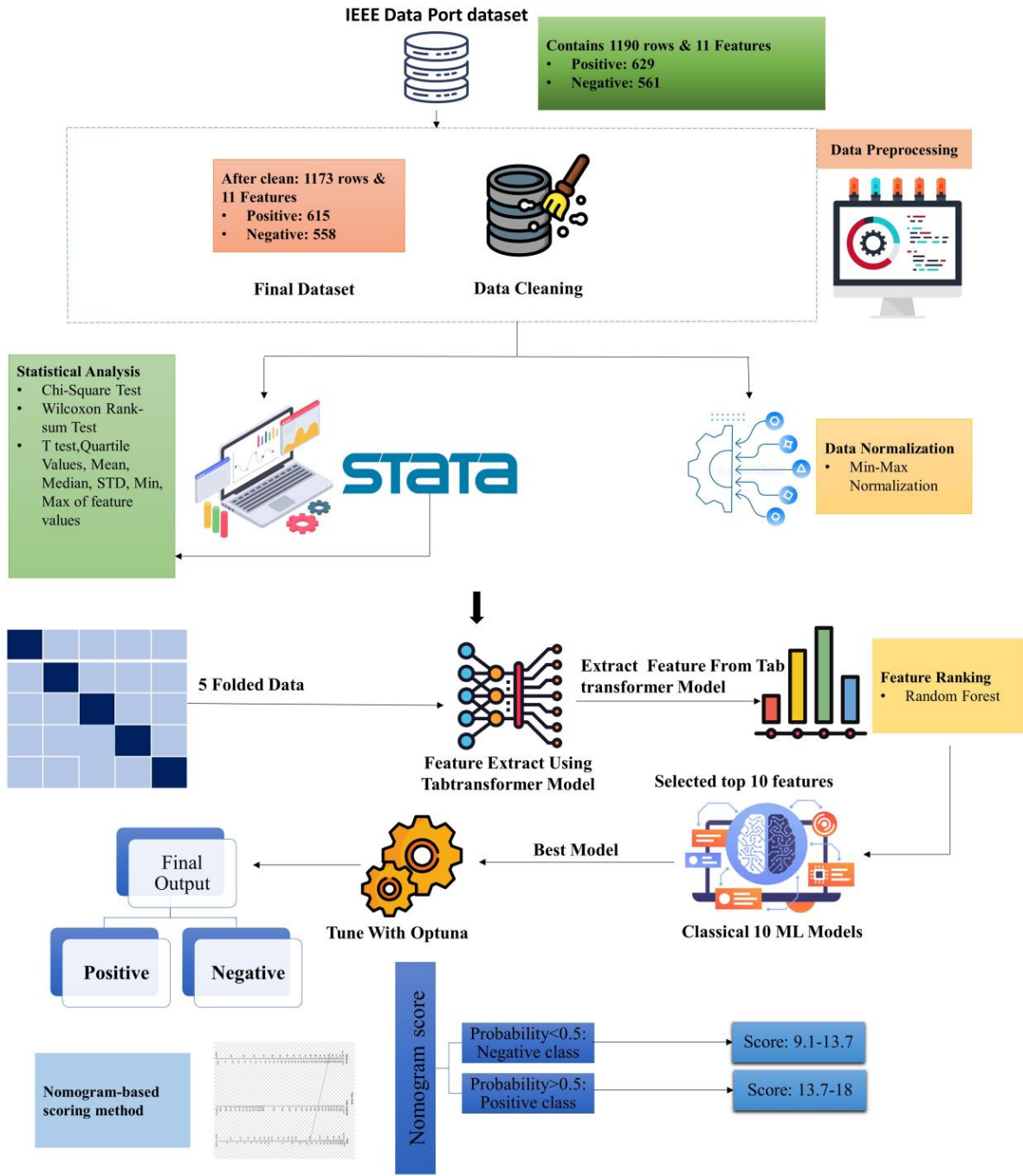

**Figure 1.** Overview of Proposed Method.

## 3. Materials and Methods

Our work aims through the implementation of CardioTabNet, a framework that utilizes transformer technology to extract superior feature spaces from clinical cardiovascular data, we intend to fundamentally alter the detection of cardiovascular disease. The overall methodology is visually depicted in **Figure 1**. The initial phase involves subjecting the dataset to thorough statistical analysis, aimed at identifying relevant characteristics. Subsequently, a meticulous data normalization process is implemented to ensure a consistent and standardized structure. Feature extraction is executed through the tab transformer model, followed by feature ranking using the RandomForest (RF) algorithm. Subsequently, a rigorous evaluation of ten ML models is conducted, leading to the identification of the top-performing model. The proposed CardioTabNet model, combining the tab transformer architecture with advanced data processing and ML techniques, effectively predicts heart disease outcomes. The model's ability to identify key features and optimize hyperparameters using Optuna (a source open-source hyperparameter optimization) highlights its potential for clinical applications in heart disease risk assessment and personalized treatment planning.

### *3.1 Data Description*

The dataset utilized in this study was sourced from the IEEE Data Port. IEEE Data Port is an online platform that provides access to various datasets related to engineering and technology fields. The dataset that we use in this paper is a comprehensive dataset for heart disease prediction that combines five well-known datasets [38-40] from different sources: Hungarian dataset, Cleveland dataset, Long Beach VA dataset, Switzerland dataset, and Statlog dataset. The dataset consists of 1190 instances with 11 features related to heart disease diagnosis. Table 2 shows the IEEE Data Port Heart Disease Prediction Dataset's full features. This dataset combines five well-known datasets from diverse sources to support heart disease prediction research. Age, gender, blood pressure, cholesterol, and exercise-induced symptoms are included. The dataset includes valuable characteristics including chest pain kinds, ECG readings, and exercise-induced angina for proper diagnosis. The goal variable—heart disease status—improves predictive modeling using the dataset. The large and multidimensional dataset from these multiple data sources allows researchers to study detailed patterns and construct superior ML models for heart disease prediction.

**Table 2:** Characteristics of the IEEE Data Port Heart Disease Prediction Dataset

| Feature | Description |
|---|---|
| **Age** | Age of the patient (in years) |
| **Sex** | Gender of the patient (0 = Female; 1 = Male) |
| **Chest Pain Type** | Type of chest pain (1 = Typical Angina; 2 = Atypical Angina; 3 = Non-Anginal Pain; 4 = Asymptomatic) |
| **Fasting Blood Sugar** | Fasting blood sugar level (>120 mg/dl; 0 = False; 1 = True) |
| **Resting ECG Results** | Resting electrocardiogram results (0 = Normal; 1 = Having ST-T wave abnormality; 2 = Showing probable or definite left ventricular hypertrophy) |
| **Cholesterol** | Cholesterol level of the patient (measured in some unit, e.g., mg/dl) |
| **Max Heart Rate** | Maximum heart rate achieved during exercise (in bpm) |
| **Exercise Induced Angina** | Presence of exercise-induced angina (0 = No; 1 = Yes) |
| **ST Depression** | ST depression induced by exercise relative to rest (in mm) |
| **Peak Exercise ST Segment Slope** | Slope of the peak exercise ST segment (1 = Upsloping; 2 = Flat; 3 = Downsloping) |
| **Presence or Absence of Heart Disease** | Target variable (0 = Absence of Heart Disease; 1 = Presence of Heart Disease) |

### *3.2 Data Preprocessing*

The initial step in the dataset preprocessing involves the application of a z-score-based filtering technique to enhance data quality by mitigating the impact of outliers[41]. Following the outlier removal process, the dataset is prepared for the training of the tab transformer model. In this subsequent phase, a systematic segregation of features is performed, distinguishing between categorical features, numerical features, and a weight column. This segmentation is outlined in **Table 3**.

**Table 3:** Summary of Numerical and Categorical Features with Weight Column.

| Feature Type | Feature Names |
|---|---|
| **Numerical** | Age |
| | RestingBP |
| | Cholesterol |
| | MaxHR |
| | Oldpeak |
| | **Weight Column Name** |
| | FastingBS |
| **Categorical** | Sex |
| | ChestPainType |
| | RestingECG |
| | ExerciseAngina |
| | ST_Slope |

We used the Synthetic Minority Over-sampling Technique (SMOTE) to enhance model robustness by addressing dataset imbalance during training. SMOTE is a data augmentation method designed specifically for handling imbalanced datasets characterized by a significant disparity in the distribution of the target variable where the minority class is underrepresented compared to the majority class [44, 45]. In addition to generating synthetic samples for the minority class, SMOTE provides a more balanced dataset, which enhances the model's ability to make accurate predictions.

*3.3 Statistical Analysis*
The dataset's characteristics were statistically interpreted using Stata/MP version 15.00. The examination encompassed statistical measures such as the mean, median, standard deviation (STD), 25th and 75th quartile values, as well as the mean and maximum values of a certain characteristic.
Additionally, p-values were adopted to determine the relationship with the output. [42].Three distinct statistical tests were conducted to derive the p-value. The Chi-square test was employed to determine the statistical correlation between the target characteristic and discrete-valued features, such as binary values. Before analysis, an assessment was made to see if the values of continuous characteristics followed a normal distribution. The t-test was employed to get the p-value if the data had a normal distribution. To determine the p-value, a Wilcoxon Rank-sum test was used when the feature values did not follow a normal distribution.

*3.4 Data Splitting*
The dataset consists of 1173 instances, each with 11 attributes after preprocessing. Implementing a 5-fold cross-validation methodology improves the resilience, dependability, and applicability of our prediction models. The dataset is divided into an 80% training set and a 20% test set to enable thorough evaluation.

*3.5 Data Normalization*
Normalizing the input data is necessary to increase the ML models' training efficiency on our data. By ensuring that each feature contributes appropriately, this normalization enhances performance as a whole. We used the Standard Scaler technique to encourage robust training and generalization. [43].

*3.6 Feature Ranking*

Feature ranking assumes a pivotal role in ML. [44], particularly in the context of datasets characterized by a considerable number of features. This step serves as a crucial precautionary measure, instrumental in addressing overfitting—an occurrence where a model excessively tailors itself to the nuances of the training data, thereby compromising its accuracy when applied to novel datasets. In our specific methodology, we employed feature ranking after feature extraction using the tab transformer Model built upon self-attention-based Transformers. Within the scope of this study, we utilized ML-based feature ranking techniques, with a specific emphasis on Random Forest.

*3.7 SMOTE Augmentation Method*
The Synthetic Minority Oversampling Technique (SMOTE) [45] is a common method for addressing class imbalance in datasets by generating synthetic samples for minority classes. It creates new samples along the line segments between an instance and its k-nearest neighbors, effectively balancing the class distribution. This reduces bias toward the majority class, enhancing model performance and reliability. By incorporating artificial data points that better represent the minority class, SMOTE ensures robust and unbiased predictions, making it a crucial step in preparing imbalanced datasets for ML. In this study, SMOTE was employed to address class imbalance in the training data. Prior to applying SMOTE, the training dataset comprised 503 positive samples and 449 negative samples, resulting in a total of 952 instances. Following the application of SMOTE, the class distribution was balanced, with the training dataset consisting of 503 positive samples and 503 synthetic negative samples, yielding a total of 1,006 instances.

*38 CardioTabNet Model Development*
The CardioTabNet model is built upon the foundational elements of the tab transformer model. [46] and classical ML models [47]. The tab transformer model, used for feature extraction, employs a tab transformer architecture based on self-attention Transformers. This architectural framework includes a column embedding layer and a series of N Transformer layers. The Transformer layers, following the principles outlined by Vaswani et al. [48], consist of a multi-head self-attention layer and a position-wise feed-forward layer.

The tab transformer operates on feature-target pairs $(x, y)$, encompassing categorical ($x_{\text{cat}}$) ad continuous ($x_{\text{cont}}$) features. Categorical features undergo embedding using a parametric technique known as Column embedding. This process involves multiple Transformer layers, each comprising a multi-head self-attention layer and a position-wise feed-forward layer. The self-attention mechanism facilitates contextual embeddings by allowing each input embedding to attend to others. These contextual embeddings, concatenated with continuous features, are then directed into an MLP for target prediction. The unique identifier approach in column embedding is elucidated as a method tailored for embedding categorical features in tabular data. Each categorical feature entails an embedding lookup table with embeddings for each class, along with an additional one for missing values. The unique identifier distinguishes classes within a column, while the embedding encompasses a category-specific segment and a feature-value-specific component. The loss function governing model training is defined as cross-entropy for classification tasks or mean square error for regression tasks. The model parameters, encompassing those for column embedding ($\varphi$), Transformer layers ($\theta$), and the top MLP layer ($\psi$), undergo learning through end-to-end training. Noteworthy is the tab transformer's unique identifier approach in column embedding, designed specifically for tabular data without positional encodings. The discussion also alludes to an ablation study comparing various embedding strategies, encompassing variations in dimensionality and the incorporation of unique identifiers.

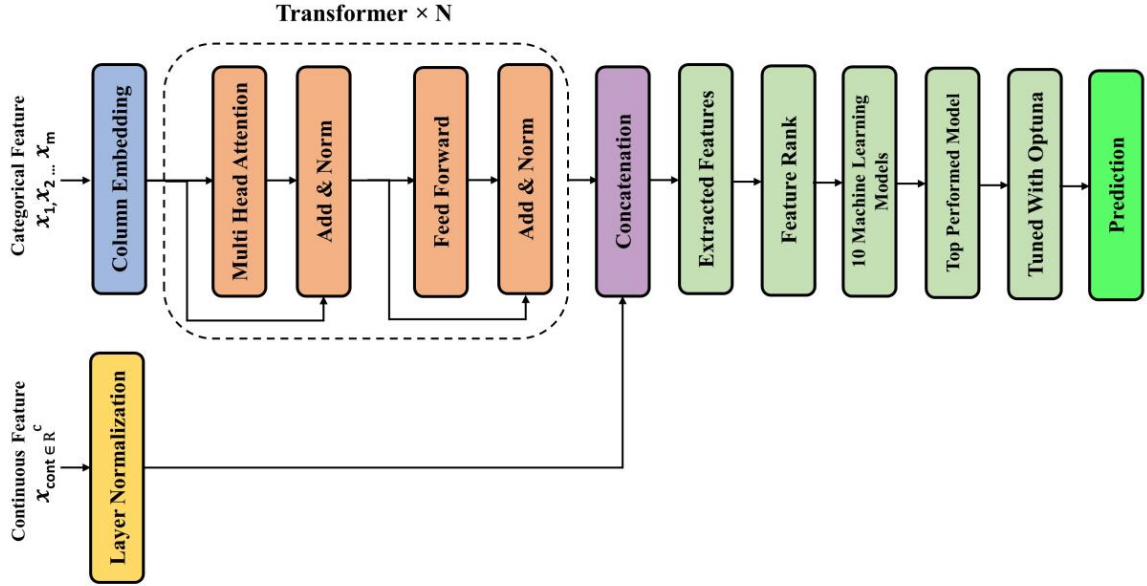

**Figure 2.** Overview of CardioTabNet Model Configuration.

**Figure 2** illustrates a systematic procedure in which categorical characteristics are subjected to column embedding and pass through a transformer block, while numerical features undergo layer normalization. Afterward, the two sets of features are combined to produce the extracted features. After performing feature extraction using tab transformer, a Random Forest (RF) method is utilized to rank the features and select the top 10. The next step entails the training of ten traditional ML models (Extra tree classifier, Random Forest Classifier, Gradient Boost Classifier, Cat Boost classifier, XGB Classifier, MLP classifier, Light Gradient Boosting Machine (LGBM) classifier, Linear Discriminant Analysis, and Logistic Regression). The model that performs the best is determined among these options. The selected model undergoes further refining through optimization using Optuna. [49], to improve its prediction capabilities. This rigorous procedure concludes with the anticipation of the ultimate result.

Let $(x, y)$ Represent a feature-target pair, where $x$ consists of categorical features $x_{cat}$ and continuous features $x_{cont}$. The continuous features $x_{cont}$ belong to $R^c$ With $c$ representing the number of continuous features. The categorical features ($x_{cat}$) are denoted by $\{x_1, x_2, \ldots, x_m\}$ where each $x_i$ Is a categorical feature for $i = 1, \ldots, m$.

Each categorical feature $x_i$ Is embedded into a parametric vector of dimension $d$ using column embedding. Let the embedding for $x_i$ be denoted by $e\phi i(xi) \in R^d$, with $\phi_i$ representing the embedding parameters for $x_i$. We denote the set of all categorical embeddings as:

$$E_\phi(x_{cat}) = \{e_{\phi_1}(x_1), \ldots, e_{\phi_m}(x_m)\} \quad (1)$$

The embeddings $E_\phi(x_{cat})$ are fed into the first Transformer layer. Each Transformer layer processes the input embeddings and outputs contextual embeddings by aggregating context from other embeddings through successive layers. This transformation is denoted by a function $f\theta$, which takes the parametric embeddings $\{e_{\phi_1}(x_1), \ldots, e_{\phi_m}(x_m)\}$ and outputs contextual embeddings $\{h_1, \ldots, h_m\}$, where each $h_i \in R^d\ for\ i = 1, \ldots, m$.

These contextual embeddings are concatenated with the continuous features $x_{\text{cont}}$ to form a vector of dimension $(d \times m + c)$. This vector is then passed to a multi-layer perceptron (MLP), represented by $g_\psi$, to predict the target $y$. The loss function $L(x, y)$, which could be cross-entropy for classification or mean squared error for regression, is minimized to learn all the parameters of tab transformer, including $\phi(column\ embeddings)$, $\theta(Transformer\ layers)$, and $\psi(MLP\ layer)$. The loss function can be written as:

$$L(x, y) = H\left(g_\psi\left(f_\theta(E_\phi(x_{\text{cat}}), x_{\text{cont}}), y\right)\right) \quad (2)$$

The Transformer consists of multi-head self-attention layers followed by position-wise feed-forward layers, with each layer having element-wise addition and normalization. The self-attention layer uses three parametric matrices: Key $(K)$, Query $(Q)$, and Value $(V)$. The embeddings are projected into these matrices to generate key, query, and value vectors. Let:

$$K \in R^{m \times k}, \quad Q \in R^{m \times k}, \quad V \in R^{m \times v} \quad (3)$$

Where $m$ is the number of embeddings inputted to the Transformer, and $k$ and $v$ are the dimensions of the key and value vectors, respectively. The attention mechanism calculates the relevance of other embeddings for each embedding using:

$$\text{Attention}(K, Q, V) = A \cdot V \quad (4)$$

Where,

$$A = \text{softmax}\left(\frac{QK^T}{\sqrt{k}}\right) \quad (5)$$

The attention matrix $A \in R^{m \times m}$ calculates how much each embedding attends to others, resulting in contextual embeddings.

For each categorical feature iii, we maintain an embedding lookup table $e_{\phi_i}(\cdot)$. If the feature has $d_i$ classes, the embedding table $e_{\phi_i}$ has $(d_i + 1)$ embeddings, where the extra embedding is used for missing values. The embedding for a value $x_i = j$ is defined as:

$$e_{\phi_i}(j) = [c_{\phi_i}, w_{\phi_{ij}}] \quad (6)$$

where,

$$c_{\phi_i} \in R^\ell, \quad w_{\phi_{ij}} \in R^{d-\ell} \quad (7)$$

Here, $\ell$ is a hyper-parameter that helps distinguish between different classes across columns, providing uniqueness.

As shown in Figure 3, pseudocode for tab transformer-based feature selection and model training algorithm presents a structured approach for training ML models for tabular data. Preprocessing is the first step in the algorithm, which ensures that the data is appropriately tokenized and cleansed. Input data is processed using the tab transformer model to identify critical patterns and derive valuable features. After the extracted features are prioritized according to their significance, the top 20 are selected for further analysis. This is followed by the initialization and training of a ML model using the selected attributes. Finally, the algorithm estimates the accuracy of the trained model using test data and provides both the model's accuracy and performance metrics. As a result of this workflow, it is possible to improve predictive performance by selecting features efficiently and training models.

```
Input:  D_{train}: Set of training data, D_{test}: Set of testing data
Output: F_{train}, F_{test}: Top N selected features, M_{ML}: Trained machine learning model

1. Preprocessing
   For each d ∈ D_{train} and d ∈ D_{test}:
     1. Clean d
     2. Normalize and prepare data
   End For
2. Feature Extraction
   Initialize F_{train} and F_{test}
   1. Extract features for D_{train} and D_{test} using TabTransformer:
      F_{train} ← TabTransformer.extract_{features}(D_{train})
      F_{test} ← TabTransformer.extract_{features}(D_{test})
3. Feature Ranking and Selection
   1. Rank features based on importance:
      Ranked_features ← FeatureRanking(F_{train}, Y_{train})
   2. Select top N features:
      F_{train_20} ← Select Top N Features (Ranked_features, F_{train}, N = 20)
      F_{test_20} ← Select Top N Features (Ranked_features, F_{test}, N = 20)
4. Model Initialization
   Initialize the machine learning model:
   M_{ML} ← InitializeMLModel()
5. Model Training:
   Train M_{ML} on F_{train_20}, Y_{train}
6. Model Evaluation:
   1. Predict Y_{test} ← M_{ML}.predict(F_{test_20})
   2. ComputeAccuracy: Accuracy = (1/|Y_{test}|) Σ_{i=1}^{|Y_{test}|} 1(Y_{test}^{(i)} = Y_{pred}^{(i)})
7. Output
   Return M_{ML}, Accuracy
```

Figure 3: Pseudocode for the tab transformer-Based Feature Selection and Model Training Algorithm.

*3.9 Nomogram-based scoring system*

Nomograms [50] are visual representations that condense statistical prediction models into a singular numerical estimation of the likelihood of an occurrence, specifically customized for an individual patient's characteristics. This leads to developing a grading system that can substantially aid healthcare practitioners in quickly differentiating between good and negative categories. Graphical interfaces that are easy for users to use are available to generate these estimates. These interfaces make it easier for healthcare professionals to use nomograms during clinical encounters, providing valuable information for clinical decision-making.

Our investigation culminates in presenting a nomogram, a tool clinicians favor for its frequent utilization during investigative processes. A nomogram was constructed in our inquiry using the most optimal features identified by the random forest feature ranking technique. In addition, calibration took place aligning it with the ground truth labels and the forecast likelihood of the occurrences. Decision Curve Analysis (DCA) was used to present a threshold for each attribute graphically. The aforementioned actions were performed with the Stata 15.

*3.10 Machine Learning Models*

In our study, we implemented feature extraction using the tab transformer model and subsequently trained ten classical ML models (Extra tree classifier, Random Forest Classifier, Gradient Boost Classifier, Cat Boost classifier, XGB Classifier, MLP classifier, Light Gradient Boosting Machine (LGBM) classifier, Linear Discriminant Analysis, and Logistic Regression) [47]. The aim was to leverage the rich representations learned by tab transformer to enhance the predictive capabilities of these traditional models.

*3.11 Evaluation Metrics*

The efficacy of the model cannot be evaluated based solely on its accuracy. To enhance the dependability of the findings, a comprehensive range of evaluation criteria was implemented, acknowledging that exclusive reliance on precision was inadequate [51]. A variety of metrics can be identified by utilizing the subsequent formulations, which span from Equation 8 to Equation 12:

$$Accuracy\ (A) = \frac{TP + TN}{TP + TN + FP + FN} \qquad (8)$$

$$Precision\ (P) = \frac{TP}{TP + FP} \qquad (9)$$

$$Recall\ (R) = \frac{TP}{TP + FN} \qquad (10)$$

$$Specificity\ (S) = \frac{TN}{TN + FP} \qquad (11)$$

$$F1 - Score\ (F1) = 2 * \frac{Precision \times Recall}{Precision + Recall} \qquad (12)$$

where True Positive, True Negative, False Positive, and False Negative are denoted, respectively, by the letters TP, TN, FP, and FN. The model performance can be obtained by evaluating the classification performance using the ROC (Receiver Operating Characteristic), AUC (area under the curve), and confusion matrix.

*3.12 Experimental Setup*
The research utilized Python 3.10 and the scikit-learn package to implement all models. Scikit-learn [52] is a prevalent Python package for ML, constructed upon NumPy, SciPy, and matplotlib. It offers a straightforward interface for executing both supervised and unsupervised learning algorithms, including classification, regression, clustering, and dimensionality reduction. The training of these models adhered to specific hardware specifications, including an Nvidia GForce 1050ti GPU, an AMD Ryzen 7 5800X 8-Core Processor, and 32GB of high RAM. We employed a straightforward pipeline for this experiment, integrated with all classical ML models.

**Table 4** shows the key hyperparameters in the tab transformer-based model, such as learning rate, weight decay, dropout rate, batch size, as well as architecture-specific parameters like transformer blocks, attention heads, and MLP hidden units. To optimize the performance of the model, these hyperparameters are crucial.

Table 4: Hyperparameter Details for the tab transformer-based Model.

| **Hyperparameter** | **Value** |
|---|---|
| Learning_Rate | 0.001 |
| Weight_Decay | 0.0001 |
| Dropout_Rate | 0.2 |
| Batch_Size | 265 |
| Num_Epochs | 500 |
| Optimizer | ADAM |
| Num_Transformer_Blocks | 3 |
| Num_Heads | 4 |
| Embedding_Dims | 8 |
| MLP_Hidden_Units_Factors | [2, 1] |
| Num_MLP_Blocks | 2 |

## 4. Numerical Results and Discussion

This section presents the outcomes derived from implementing the algorithms described in the methodology section. In addition to presenting the data, a thorough and analytical discussion has been carried out to justify the outcome.

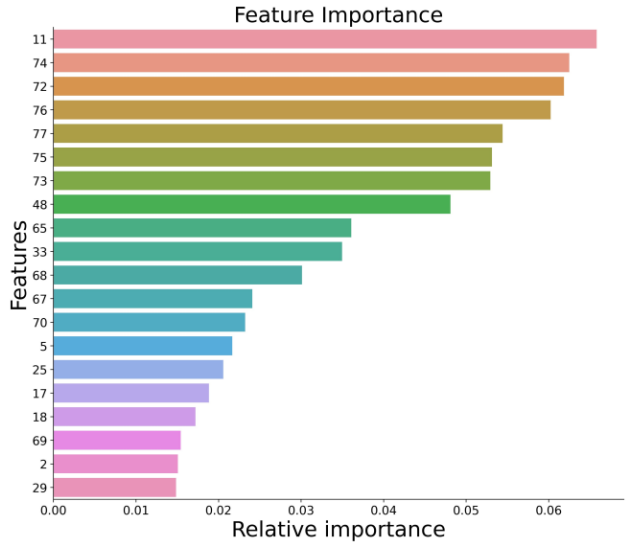

**Figure 4:** Top-20 extracted features ranked using Random Forest feature selection algorithm.

### 4.1 Feature Ranking

The CardioTabNet model was employed to extract features, leveraging the capabilities of the tab transformer. Subsequently, a Random Forest Model was utilized to rank the extracted features. In comparison to alternative models such as XGBoost and ExtraTree, the Random Forest model demonstrated superior performance. The selection process focused on identifying the top 10 features, which were then utilized for the final prediction. **Figure 4** visually presents the top 20 features extracted through the tab transformer, providing a comprehensive illustration of the key contributors to the predictive modeling process.

### 4.2 Statistical Analysis

**Table 5** shows the statistical analysis of the dataset's characteristics. P-value indicates deviation of two observations which for our case is the study between the target variable and the corresponding feature column. The lower p-value would clearly propose that differences between the two columns is not due to random event but prove a strong association between the two features. Therefore, a smaller value of 0.05 or less value of p would provide stronger significance. The succeeding table sheds some light on to the matter of statistical association between features and the output label. Apart from rest_ecg_left ventricular hypertrophy and st_slope_normal, all other biomarkers carry a value less than 0.05. The male population is significantly affected by the disease, accounting for around 89.11% of the positive cases. The study included individuals aged 28 to 77, with a higher percentage of older patients in the disease-positive group.

**Table 5:** Statistical Analysis on IEEE Port Heart disease dataset.

| S.No | Feature name | Positive class | Negative class | Total | Test | Test Statistic | P-value |
|---|---|---|---|---|---|---|---|

| # | Variable | | | | Test | Statistic | p-value |
|---|---|---|---|---|---|---|---|
| 1 | Gender | | | | Chi-square test | 114.7073 | <0.001 |
| | • Male (%) | 548(89.11%) | 349(62.54%) | 897(76.47%) | | | |
| | • Female (%) | 67(10.89%) | 209(37.45%) | 276(23.53%) | | | |
| 2 | Age | | | | T-test | -9.4233 | <0.001 |
| | • Mean ± STD | 56.07±8.63 | 51.09±9.46 | 53.7±9.37 | | | |
| | • Median | 57.00 | 51.00 | 54.00 | | | |
| | • Q1, Q3 | 51, 62 | 44, 57 | 47, 60 | | | |
| | • Min, Max | 31, 77 | 28, 76 | 28, 77 | | | |
| 3 | resting_blood_pressure | | | | Rank-sum test | -3.884 | <0.001 |
| | • Mean ± STD | 133.67±17.73 | 129.74±16.31 | 131.8±17.17 | | | |
| | • Median | 131.00 | 130.00 | 130 | | | |
| | • Q1, Q3 | 120, 144 | 120, 140 | 120, 140 | | | |
| | • Min, Max | 92, 185 | 80, 180 | 80, 185 | | | |
| 4 | Cholesterol | | | | T-test | 7.097 | <0.001 |
| | • Mean ± STD | 190.4±116.86 | 230.45±67.31 | 209.45±98.53 | | | |
| | • Median | 226.00 | 231.00 | 229.00 | | | |
| | • Q1, Q3 | 135, 271 | 201, 267 | 188, 269 | | | |
| | • Min, Max | 0, 491 | 0, 468 | 0, 491 | | | |
| 5 | fasting_blood_sugar | | | | Chi-square test | 54.11 | <0.001 |
| | • True (%) | 182(29.59%) | 67(12.01%) | 249(21.23%) | | | |
| | • False (%) | 433(70.41%) | 491(87.99%) | 924(78.77%) | | | |
| 6 | max_heart_rate_achieved | | | | T-test | 15.78 | <0.001 |
| | • Mean ± STD | 129.77±23.20 | 150.94±22.68 | 139.84±25.27 | | | |
| | • Median | 128.00 | 153.50 | 141.00 | | | |
| | • Q1, Q3 | 113, 147 | 137.25, 169.00 | 121, 160 | | | |
| | • Min, Max | 67, 182 | 69, 202 | 67, 202 | | | |
| 7 | exercise_induced_angina | | | | Chi-square test | 274.26 | <0.05 |
| | • Yes (%) | 376(61.14%) | 78(13.98%) | 454(38.70%) | | | |
| | • No (%) | 239(38.86%) | 480(86.02%) | 719(61.30%) | | | |
| 8 | st_depression | | | | Rank-sum test | -14.152 | <0.001 |
| | • Mean ± STD | 1.34±1.18 | 0.46±0.73 | 0.92±1.083 | | | |
| | • Median | 1.20 | 0.00 | 0.60 | | | |
| | • Q1, Q3 | 0.1, 2 | 0, 0.8 | 0, 1.6 | | | |
| | • Min, Max | -2.60, 6.2 | -1.1, 4.2 | -2.60, 6.2 | | | |
| 9 | chest_pain_type_atypical angina | | | | Chi-square test | 157.21 | <0.001 |
| | • Yes (%) | 29(4.72%) | 184(32.97%) | 213(18.16%) | | | |
| | • No (%) | 586(95.28%) | 374(67.03%) | 960(81.84%) | | | |
| 10 | chest_pain_type_non-anginal pain | | | | Chi-square test | 65.25 | |
| | • Yes (%) | 87(14.15%) | 191(34.23%) | 278(23.70%) | | | |
| | • No (%) | 528(85.85%) | 367(65.77%) | 895(76.30%) | | | <0.001 |
| 11 | chest_pain_type_typical angina<br>• Yes (%)<br>• No (%) | 25(4.07%)<br>590(95.93%) | 41(7.35%)<br>517(92.65%) | 66(5.63%)<br>1107(94.37%) | Chi-square test | 5.94 | <0.05 |
| 12 | rest_ecg_left ventricular hypertrophy<br>• Yes (%)<br>• No (%) | 174(28.29%)<br>441(71.71%) | 144(25.81%)<br>414(74.19%) | 318(27.11%)<br>855(72.89%) | Chi-square test | 0.9152 | 0.34 |
| 13 | rest_ecg_normal<br>• Yes (%)<br>• No (%) | 323(52.52%)<br>292(47.48%) | 352(63.08%)<br>206(36.92%) | 675(57.54%)<br>498(42.46%) | Chi-square test | 13.36 | <0.001 |
| 14 | st_slope_flat<br>• Yes (%)<br>• No (%) | 451(73.33%)<br>164(26.67%) | 121(21.68%)<br>437(78.31%) | 572(48.76%)<br>601(51.24%) | Chi-square test | 312.36 | <0.001 |
| 15 | st_slope_normal<br>• Yes (%)<br>• No (%) | 1(0.16%)<br>614(99.84%) | 0(0%)<br>558(100%) | 1(0.09%)<br>1172(99.91%) | Chi-square test | 0.91 | 0.34 |
| 16 | st_slope_upsloping<br>• Yes (%)<br>• No (%) | 107(17.40%)<br>508(82.60%) | 415(74.37%)<br>143(25.63%) | 522(44.50%)<br>651(55.50%) | Chi-square test | 384.52 | <0.001 |
| 17 | Target (%) | 615(52.43%) | 558(47.57%) | 1173 | | | |

Q1 = First quarter, Q3 = Third quarter.

### *4.3 Classification using CardioTabNet Model*

In Table 6, the results of the first feature extraction are presented, along with their accuracy for each fold. As part of the initial analysis, features were extracted using the tab transformer model from a dataset divided into five folds. By applying the Random Forest algorithm, a comprehensive feature ranking process was conducted, which enabled the identification and selection of the top 10 features. Based on the selected top 10 features, an ensemble of 10 ML models was trained.

**Table 6:** Result of Feature extraction with tab transformer.

| Fold | Accuracy (%) |
|---|---|
| 1 | 89.74 |
| 2 | 90.12 |
| 3 | 88.95 |
| 4 | 89.30 |
| 5 | 90.04 |

As a result of feature extraction using the tab transformer, **Table 7** presents the top 10 ML models. ExtraTree Classifier achieved the highest accuracy of 92.82%, with notable precision, recall, and F1 score values of 91.04%, 95.18%, and 93.02%, respectively. Additionally, the RandomForestClassifier achieved robust precision, recall, and F1 score metrics, achieving an accuracy of 92.18%. CatBoost_untuned and GradientBoostingClassifier both produced competitive results, showing their effectiveness in classifying data. Both the XGBClassifier and AdaBoostClassifier performed well in terms of precision and recall, though they displayed slightly lower accuracy. As a result of the MLPClassifier and LGBM models, consistent results were obtained, supporting their reliability. The results of LinearDiscriminantAnalysis and LogisticRegression were notable for their relatively low sensitivity (recall) when compared with other classifiers despite achieving good accuracy.

Table 7: Result for the Top-10 ML After Feature extraction with tab transformer.

| Model | Accuracy | Precision | Recall | F1 Score | Specificity | AUC |
|---|---|---|---|---|---|---|
| **ExtraTree Classifier** | **92.82%** | **91.04%** | **95.18%** | **93.01%** | **90.58%** | **92.88%** |
| **Random Forest Classifier** | 92.17% | 89.77% | 95.41% | 92.46% | 88.90% | 92.16% |
| **Gradient Boosting Classifier** | 91.92% | 90.30% | 94.14% | 92.13% | 89.76% | 91.95% |
| **CatBoost** | 90.00% | 88.05% | 92.88% | 90.32% | 87.17% | 90.03% |
| **XGB Classifier** | 86.02% | 84.45% | 89.30% | 86.57% | 83.02% | 86.16% |
| **AdaBoost Classifier** | 84.10% | 83.15% | 86.77% | 84.67% | 81.79% | 84.28% |
| **MLP Classifier** | 84.10% | 81.75% | 88.31% | 84.83% | 79.76% | 84.03% |
| **LGBM** | 83.84% | 81.93% | 88.08% | 84.62% | 79.92% | 84.00% |
| **Linear Discriminant Analysis** | 82.43% | 81.95% | 84.02% | 82.89% | 81.02% | 82.52% |
| **Logistic Regression** | 82.43% | 81.96% | 83.98% | 82.90% | 80.98% | 82.48% |

*4.3.1 HyperParameter Tune with Optuna*

Optuna, a Bayesian optimization library, efficiently tunes hyperparameters for ML models [49], This approach significantly improves model performance, demonstrating Optuna's potential as a powerful hyperparameter tuning tool. A meticulous hyperparameter tuning process was conducted utilizing Optuna to optimize the parameters of the ExtraTree Classifier, resulting in the identification of the best-performing model configuration. The optimal hyperparameters, determined through the hyperparameter tuning process with Optuna, were as follows: 'n_estimators' set to 176, 'max_depth' set to 19, and 'bootstrap' set to True.

**Table 8** showcases the five-fold cross-validated results for the ExtraTree Classifier after hyperparameter tuning with Optuna, which was identified as the best-performing model initially. Across all folds, the tuned ExtraTree model consistently demonstrated high accuracy, precision, recall, and F1 score values, affirming its robust performance in the classification task. Notably, the model achieved an average accuracy **of 94.089%,** with a balanced and elevated performance across all evaluated metrics. These results underscore the efficacy of the hyperparameter tuning process in enhancing the model's overall predictive capability and generalizability.

Table 8: Results on extra-tree classifier after tuning with the Optuna.

| Fold | Model | Accuracy | Precision | Recall | F1 Score | Specificity | AUC |
|---|---|---|---|---|---|---|---|
| 1 | ExtraTrees | 94.23% | 92.77% | 96.25% | 94.47% | 92.10% | 94.95% |
| 2 | | 97.43% | 97.64% | 97.65% | 97.64% | 97.18% | 98.41% |
| 3 | | 90.38% | 84.14% | 97.18% | 90.19% | 84.70% | 90.95% |
| 4 | | 92.90% | 90.69% | 96.3% | 93.41% | 89.18% | 95.11% |

| 5 | (Tuned With Optuna) | 95.48% | 93.90% | 97.47% | 95.65% | 93.42% | 97.63% |
|---|---|---|---|---|---|---|---|
| **Average** | | **94.08%** | **92.84%** | **97.37%** | **94.47%** | **91.92%** | **95.01%** |

The Receiver Operating Characteristic (ROC) curve, presented in **Figure 5(a)**, illustrates the performance of the tuned ExtraTree model across five-fold data in the context of heart disease classification. The Area Under the Curve (AUC) serves as a comprehensive metric, representing the model's overall performance by averaging the True Positive Rate (TPR) and False Positive Rate (FPR) across various threshold settings. The AUC score of 0.95 for the tuned ExtraTree Model signifies its excellence in distinguishing between positive and negative cases, further affirming its efficacy in the heart disease detection task.

The confusion matrix and associated metrics for the Tuned ExtraTree Model are visually depicted in **Figure 5(b)**. The illustration underscores the model's exemplary performance in the classification task, demonstrating high accuracy, precision, recall, and F1 score values. These metrics collectively affirm the model's capability to accurately classify both positive and negative cases, underscoring its proficiency in the task at hand.

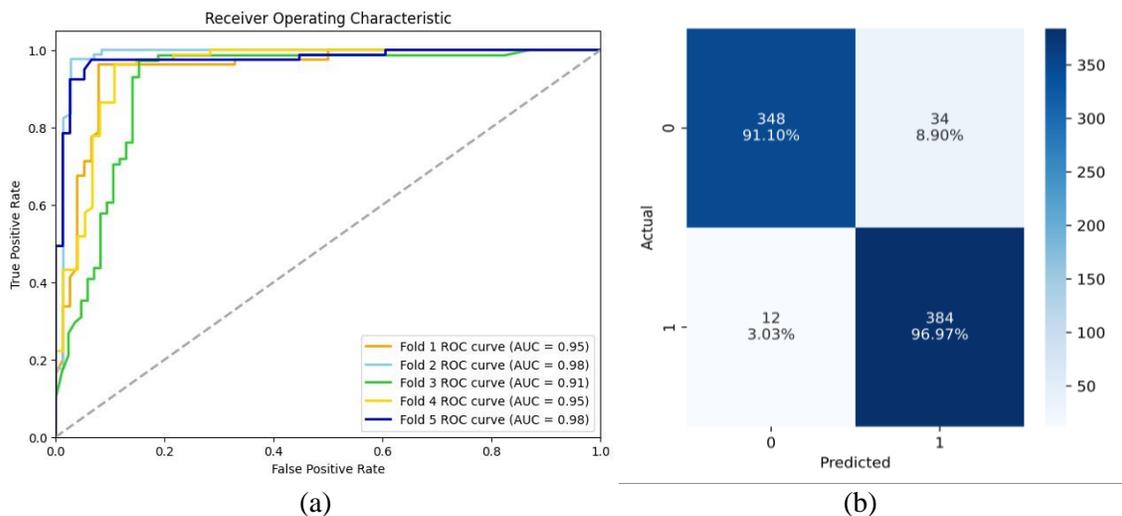

**Figure 5:** (a) ROC curve of each fold for Tuned ExtraTree Classifier, (b) Confusion Matrix for Tuned Extra Tree Classifier.

*4.4 Nomogram-based scoring technique*

In this work, we utilized regression analysis to construct a predictive outcome model. This model was then visualized using a multivariate logistic regression nomogram. The statistical pipeline produced a linear result calculated by applying the 10 most significant characteristics received from the feature extraction process described in previous sections. In the supplementary table Table 1S, a comprehensive regression analysis that led to the development of the nomogram is detailed. The z-value is a commonly used statistic that specifies the position of a result inside the nomogram. As the frequency of the z-value shifts in favor of a higher score rises, the importance of the corresponding independent feature becomes more visible in the nomogram. In that sense, feature_10 was removed in the nomogram as the values in the table suggested the feature being of lowest contributor. Feature 7 was automatically omitted in the regression analysis for being collinear.

The nomogram of Figure 6 illustrates that Feature 5 exhibits the greatest degree of dispersion, which corresponds to the elevated score indicated in both the table and the nomogram. The nomogram of this project is constructed using characteristics obtained from the extraction procedure. It represents the outcome of each event and calculates the probability result on the bottom scale. The total score displayed below the 50% likelihood score helps identify the negative classifications in heart disease categorization. Accordingly, we may assert that a score ranging from 9.1 to 13.7 in the scoring method indicates a sufficiently high likelihood of the categorization being negative. Conversely, scores ranging from 13.7 to 18 are considered part of the positive class since the chance is higher than 50%.

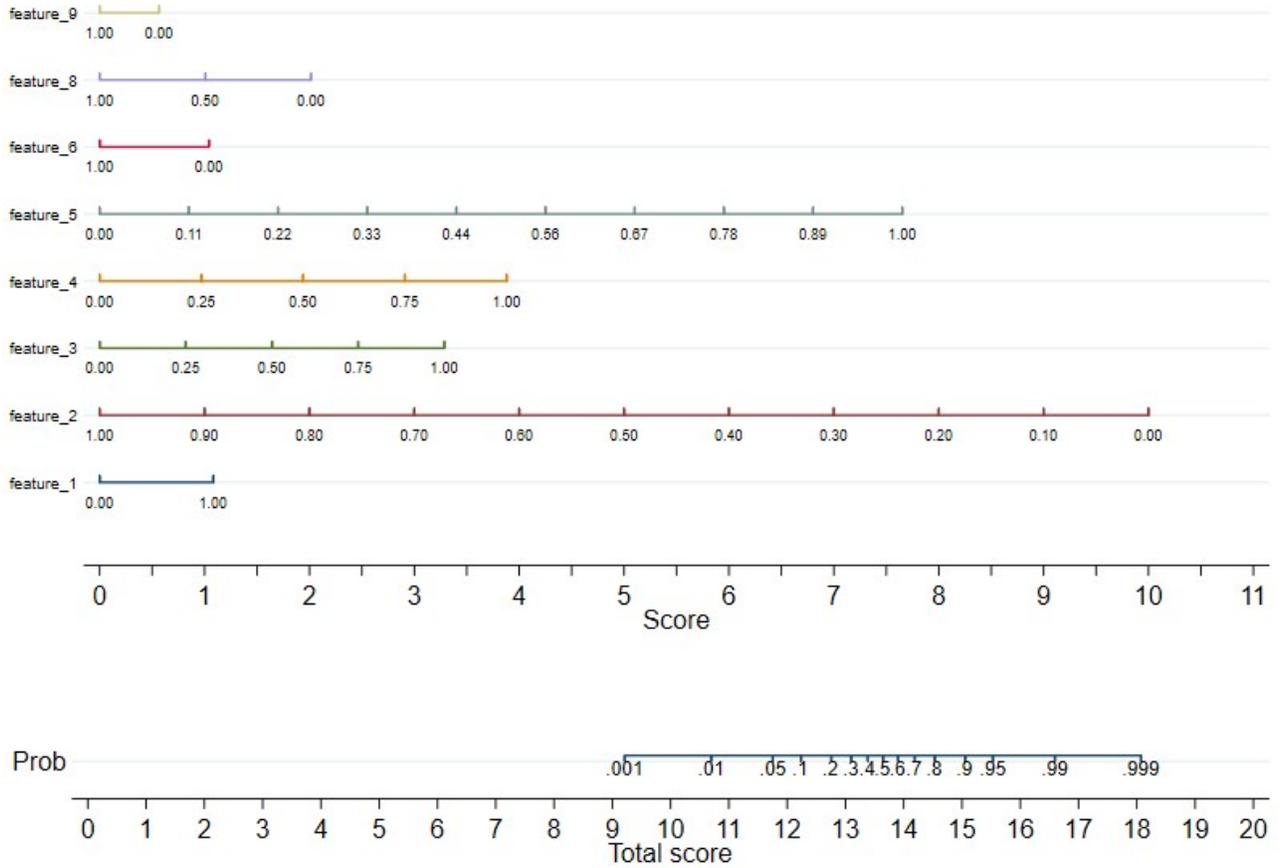

**Figure 6:** Nomogram for predicting outcome probability.

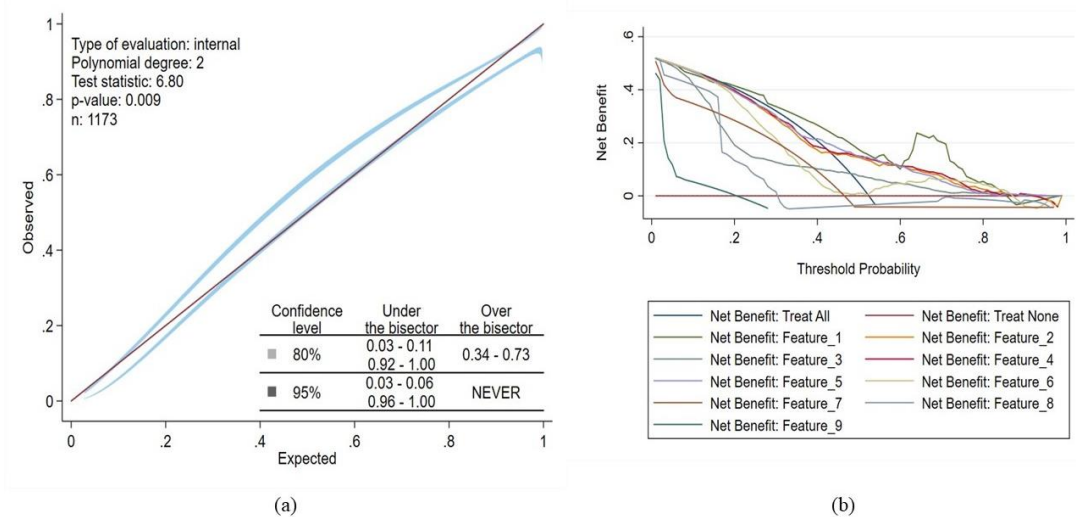

**Figure 7:** (a) Calibration plot, (b) decision curve analysis.

This work demonstrates the use of internal validation to develop a calibration technique for analyzing and measuring the quality of logistic regression's probability estimations in predicting the actual output labels shown in Figure 7 (a). A significant amount of the calibration belt is located on the diagonal bisector, showing an equal distribution between the areas above and below it. Therefore, the anticipated probabilities were perfectly aligned with the actual results. Figure 6 (b) illustrates the total advantages that each feature included in the study contributes to the decision-making process for the model in identifying positive and negative classes.

*4.5 Discussion*
Heart disease, or cardiovascular disease, encompasses a range of disorders that impact the heart and blood arteries. It is a prominent factor contributing to illness and death on a global scale, affecting a significant number of individuals. Cardiovascular disorders, which can be classified into four kinds including coronary heart disease, heart failure, congenital heart disease, and cardiomyopathy, provide a substantial risk [53]. Timely and accurate detection of cardiac disease is crucial to prevent worsening and save lives. Early detection and prompt intervention in cardiovascular diseases play a critical role in reducing premature mortality [54]. Predictive models are essential tools for identifying individuals at risk, enabling proactive healthcare interventions. Recognizing these diseases early is crucial for effective intervention and improved overall health outcomes. ML presents a transformative approach to addressing the complex challenge of cardiovascular diseases, offering innovative solutions for early detection, risk prediction, and personalized healthcare interventions [19-26]..

In this study, we propose a novel model, CardioTabNet, for cardiovascular disease prediction. The use of Transformers, originally designed for natural language processing tasks [55, 56], in tabular data represents a significant advancement in machine learning and predictive modeling. Now, the transformers are also used in medical image classification applications [60]. Transformers, particularly those that rely on self-attention mechanisms, have exhibited exceptional efficacy in discerning complex relationships present in structured and sequential data [47]. Self-attention mechanisms enable transformers to evaluate the relative significance of individual features to others. This approach has the potential to yield a more intricate comprehension of feature interactions in contrast to conventional models that might fail to adequately capture such interactions. It can process unprocessed data more efficiently and necessitates less manual feature engineering in comparison to conventional models, owing to their sophisticated architecture. This can yield a substantial benefit by decreasing the amount of time and effort needed to preprocess the data. Consequently, they have also been found to apply to tabular datasets. Transformers are excellent at identifying long-range data dependencies [48]. Transformers are successful in capturing the relationships between distant categorical features in the context of tabular data [48], where features may have intricate interactions and dependencies.

The initial phase involves a comprehensive statistical analysis of the dataset using Stata/MP version 15.00. Statistical measures such as mean, median, standard deviation (STD), 25th and 75th quartile values, as well as the mean and maximum values, were employed for a thorough characterization of the dataset which was presented in **Table 4**. The CardioTabNet model leverages tab transformer for feature extraction, which is based on self-attention transformers. The transformer layers transform categorical feature embeddings into robust contextual embeddings. Tab transformers possess the capability to accommodate various feature types, encompassing both categorical and numeric values. The self-attention mechanism facilitates the model's ability to seamlessly process mixed-type data by permitting it to allocate various levels of importance to distinct features. Particularly advantageous in real-world datasets where diverse feature types are prevalent is this adaptability. Following feature extraction, RandomForest is employed for feature ranking, a pivotal step in addressing overfitting and ensuring model accuracy on novel datasets. The top 10 extracted features are then utilized to train 10 ML models, with detailed results presented in **Table 5**. The GradientBoostingClassifier and CatBoost models demonstrated competitive results, emphasizing their effectiveness in the classification task. Although the XGBClassifier and AdaBoostClassifier displayed slightly lower accuracy, they maintained respectable precision and recall values, contributing to their overall reliability. Consistent and reliable outcomes were observed with the MLPClassifier and LGBM models.The ExtraTree Classifier emerges as a standout performer with an accuracy of 92.82%, accompanied by noteworthy precision, recall, and F1 score values of 91.04%, 95.18%, and 93.02%, respectively. Subsequent tuning of the ExtraTree model using Optuna further enhances its performance, achieving an average accuracy of 94.08% across all evaluated metrics. The analysis extends to the examination of the ROC curve, revealing an AUC score of 95.01% for the tuned ExtraTree Model. This underscores its exceptional ability to distinguish between positive and negative cases, reinforcing its efficacy in heart disease detection.

Finally, logistic regression and nomogram analysis are employed to enhance binary classification accuracy and facilitate differentiation between positive and negative classes. Regression analysis, utilizing the 10 most significant features obtained from the extraction process, culminates in the construction of a predictive outcome model visualized through a multivariate logistic regression nomogram. This comprehensive statistical pipeline contributes to a nuanced understanding of cardiovascular disease prediction, emphasizing both accuracy and interpretability.

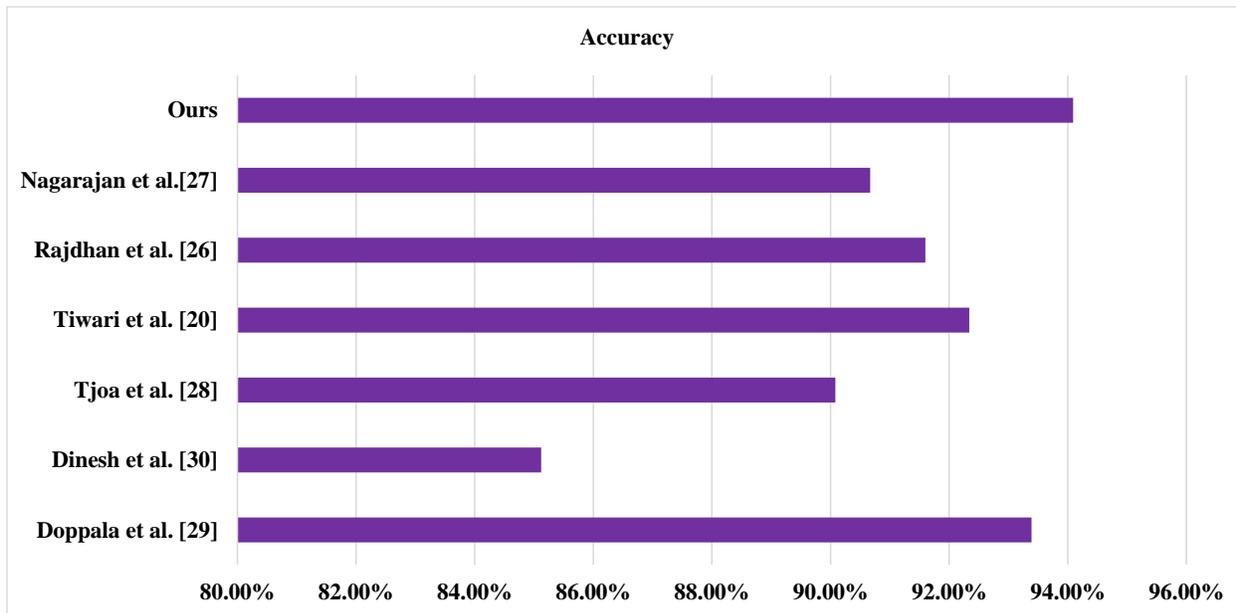

**Figure 8:** Comparative Analysis with Previous Studies.

A comparative analysis is presented in **Figure 8** alongside other works in the IEEE Port Heart Disease dataset. Our CardioTabNet model attained an accuracy of 94.09%, surpassing the accuracies reported by previous studies. Specifically, Doppala et al. achieved an accuracy of 93.39%, Dinesh et al. reported 85.12%, Tjoa et al. achieved 90.08%, Tiwari et al. reported 92.34%, Rajdhan et al. achieved 91.60%, and Nagarajan et al. reported 90.67%. Comparing these results, it's evident that our CardioTabNet model outperforms previous studies in terms of accuracy. This highlights the effectiveness of our model in accurately predicting cardiovascular diseases. The superior performance of CardioTabNet underscores its potential as an advanced tool for cardiovascular disease prediction, offering enhanced accuracy and reliability compared to existing approaches.

For future studies, tab transformer can be applied to different biological data. The integration of attention mechanisms, especially in models like tab transformer, enhances the capacity to capture hidden features and interactions in biological data and networks. This application minimizes the reliance on extensive feature engineering and effectively addresses the inherent ambiguity in biological datasets. Attention mechanisms, adept at handling noisy and conflicting information, dynamically prioritizing features, making the model robust to complex biological scenarios. This approach holds promise for future applications across diverse biological data types, offering a streamlined and adaptable method for meaningful feature extraction and interpretation.

*4.6 Limitations and Future Work*

CardioTabNet, a novel model for the prediction of cardiovascular diseases, is presented in this study. Based on clinical cardiovascular data, the model employs transformer technology to extract a high-quality feature space. In structured and sequential data, transformers, particularly those that use self-attention mechanisms, have proven highly effective at identifying complex patterns. This study used data from the IEEE Data Port, which provides access to a variety of engineering and technology datasets. We used a comprehensive heart disease dataset that combines five well-known datasets: the Hungarian dataset, the Cleveland dataset, the Long Beach VA dataset, the Switzerland dataset, and the Statlog dataset. A total of 1,190 instances with 11 features relevant to heart disease diagnosis make up this combined dataset.

However, the relatively modest size of the dataset is one of the limitations of this study. Despite the variety of data, the dataset does not fit the definition of a large-scale dataset, which may make it difficult for the model to generalize to a broader and more heterogeneous population. We plan to address this limitation by expanding our dataset to include data from larger and more diverse populations in the future. As part of our next objective, we will collect a custom dataset tailored specifically for cardiovascular diseases, which will lead to a more comprehensive model training and enhance the model's generalization capabilities across a variety of demographic and clinical settings. As a result, CardioTabNet will be further validated in real-world scenarios for robustness and effectiveness.

## 5. Conclusion

In conclusion, this study introduces CardioTabNet, a novel model designed for the prediction of cardiovascular diseases. The model incorporates transformer technology to extract a high-quality feature space, employing a feature ranking strategy based on clinical cardiovascular data. Transformers, particularly those leveraging self-attention mechanisms, have demonstrated remarkable effectiveness in identifying intricate links within structured and sequential data. The methodology involves a multi-step approach: initial feature extraction utilizing the tab transformer model, followed by training with ten classical ML classifiers. Our comprehensive statistical analysis demonstrates the efficacy of CardioTabNet, particularly when paired with the tuned ExtraTree Classifier, which exhibits outstanding performance. The model achieves an impressive average accuracy of 94.08%. Noteworthy precision, recall, and F1 score values further support its effectiveness, standing at 92.84%, 97.37%, and 94.47%, respectively. This remarkable performance underscores the effectiveness of the tuning process, emphasizing the model's ability to accurately classify both positive and negative cases in the context of cardiovascular disease detection. The overall evaluation metrics, including specificity and AUC, further contribute to an impressive average performance of 95.014%. Looking forward, future work could explore further refinement of the model through advanced optimization techniques and integration of additional clinical data. Additionally, expanding the model's applicability to diverse datasets and collaborating with healthcare professionals could enhance its real-world effectiveness in early-stage cardiovascular disease detection.

**Institutional Review Board Statement:** Not applicable

**Funding:** This work was made possible by student grant# QUST-1-CENG-2024-1723 from Qatar University. The statements made herein are solely the responsibility of the authors. The open-access publication cost is covered by Qatar National Library.

**Informed Consent Statement:** Not applicable

**Data Availability Statement:** The processed dataset used in this study can be made available upon a reasonable request to the corresponding author.

**Conflicts of Interest:** Authors have no conflict of interest to declare.